\documentclass[10pt,twocolumn,letterpaper]{article}

\usepackage[pagenumbers]{wacv} % To force page numbers, e.g. for an arXiv version

\usepackage{booktabs}
\usepackage{subcaption}
\usepackage{graphicx}
\usepackage{dblfloatfix}

\definecolor{wacvblue}{rgb}{0.21,0.49,0.74}
\usepackage[pagebackref,breaklinks,colorlinks,allcolors=wacvblue]{hyperref}

\def\wacvPaperID{1660} % *** Enter the WACV Paper ID here
\def\confName{WACV}
\def\confYear{2027}

\title{Multimodal Scenario Similarity Search for Autonomous Driving}

\author{Tamás Matuszka\\
aiMotive\\
{\tt\small tamas.matuszka@aimotive.com}
\and
András Tamásy\\
aiMotive\\
{\tt\small andras.tamasy@aimotive.com}
\and
Balázs Szolár\\
aiMotive\\
{\tt\small balazs.szolar@aimotive.com}
}

\begin{document}
\maketitle
\begin{abstract}
Large-scale autonomous-driving datasets contain vast numbers of recorded scenarios, creating a need for efficient retrieval methods that can identify situations similar to a given query. Existing approaches typically rely on either visual representations or motion-based descriptions, making it difficult to understand their relative strengths and limitations for scenario retrieval. In this work, we present a multimodal framework for autonomous-driving scenario retrieval that combines visual and trajectory-based representations within a unified retrieval pipeline. We investigate two trajectory-based approaches: Exo-Trajectory, an explicit matching method based on surrounding-agent motion, and ScenarioFormer, a transformer-based representation learned from object trajectories using contrastive learning. We compare these approaches against strong vision-based baselines and analyze their behavior across a diverse set of driving scenarios. Experimental results show that trajectory representations provide strong retrieval performance for motion-centric events such as cut-ins, turning maneuvers, and traffic queueing, while visual embeddings excel when appearance cues are informative. Most importantly, combining visual and trajectory information consistently improves retrieval quality, yielding the best overall performance. These findings demonstrate that appearance and motion capture are complementary notions of scenario similarity and motivate multimodal retrieval systems for autonomous-driving data mining, dataset curation, and scenario-based validation.
\end{abstract}
    
\section{Introduction}
\label{sec:intro}

Modern autonomous-driving systems generate and process enormous amounts of sensor data, resulting in datasets containing millions of recorded driving scenarios. Efficient retrieval of scenarios similar to a given query is essential for dataset curation, corner-case discovery, validation, and data-driven development workflows. However, defining similarity between driving scenarios remains a challenging problem. Two scenarios may appear visually different while exhibiting similar traffic interactions, or conversely share a similar visual context while containing fundamentally different vehicle and agent behaviors.

Existing scenario retrieval approaches typically focus on a single notion of similarity. Vision-based methods leverage video representations to capture scene appearance and spatio-temporal context, while trajectory-based approaches compare vehicle motions and traffic participant behavior. Although both perspectives are relevant for autonomous-driving applications, their relative strengths and limitations have not been systematically studied within a unified retrieval framework.

In this work, we investigate multimodal scenario retrieval by combining visual and trajectory-based representations. We evaluate two vision-based retrieval methods and introduce two trajectory-based approaches. The first, Exo-Trajectory, explicitly compares the motion patterns of surrounding traffic participants using trajectory matching. The second, ScenarioFormer, is a transformer-based representation learned from object trajectories using contrastive learning. Together, these methods enable a direct comparison between appearance-driven and motion-driven notions of scenario similarity.

Experimental results on a manually annotated driving-scene similarity benchmark show that visual and trajectory representations capture complementary information. Vision models perform best when appearance cues are informative, whereas trajectory-based methods excel on motion-centric scenarios such as cut-ins, turning maneuvers, and traffic queueing. Furthermore, combining visual and trajectory information consistently improves retrieval quality and achieves the best overall performance.

The main contributions of this work are:

\begin{itemize}
\item We present a multimodal framework for autonomous-driving scenario retrieval that enables the joint study of visual and trajectory-based similarity representations.

\item We introduce two trajectory-based retrieval approaches: Exo-Trajectory, an explicit matching method based on surrounding-agent motion, and ScenarioFormer, a transformer-based trajectory representation learned using contrastive learning.

\item We demonstrate through extensive experiments that visual and trajectory representations capture complementary notions of scenario similarity and that their fusion yields the strongest retrieval performance.
\end{itemize}

\section{Related Work}
\label{sec:related_work}

\subsection{Visual Scenario Retrieval}

Recent advances in video representation learning have enabled effective retrieval and similarity search in large-scale video collections. Transformer-based architectures such as ViViT~\cite{vivit} learn spatio-temporal representations directly from video sequences and have achieved strong performance across a variety of video understanding tasks. More recently, large vision-language models such as Qwen3-VL~\cite{qwen} have demonstrated impressive transfer capabilities, producing semantically meaningful embeddings that can be used for retrieval without task-specific supervision. Such representations capture scene appearance, object context, and high-level semantics, making them attractive for autonomous-driving scenario search. However, visual representations may struggle to distinguish scenarios that exhibit similar appearance while differing significantly in agent behavior and motion patterns.

\subsection{Trajectory-Based Scenario Retrieval}

Trajectory information provides an alternative notion of similarity by explicitly modeling the motion of traffic participants. Traditional approaches compare trajectories using geometric distance measures such as Dynamic Time Warping (DTW)~\cite{dtw} or Fréchet distance~\cite{frechet}, which have been widely used for trajectory matching and motion analysis. In autonomous driving, trajectory-based retrieval has been explored for scenario mining, corner-case discovery, and dataset search, where the objective is to identify situations exhibiting similar interactions between vehicles and other road users. Compared with visual representations, trajectory-based methods focus directly on dynamic behavior and are therefore particularly suitable for motion-centric scenarios.

\subsection{Trajectory Representation Learning}

Recent work has increasingly adopted learned trajectory representations to capture higher-level behavioral semantics. Transformer architectures have proven effective for modeling interactions among multiple traffic participants and forecasting future motion. Scene Transformer~\cite{sceneformer} introduced a unified attention-based framework for representing complex traffic scenes, demonstrating the ability of transformer models to encode rich relational information between agents. Contrastive learning has further emerged as a powerful approach for learning task-agnostic embeddings from structured data by encouraging similar samples to be mapped close together in the representation space. Motivated by these advances, our proposed ScenarioFormer learns trajectory embeddings directly from agent trajectories using a transformer encoder trained with a contrastive objective, enabling efficient retrieval based on scene dynamics.

\section{Methodology}
\label{sec:method}

We investigate autonomous-driving scenario retrieval using two complementary modalities: visual representations extracted from video data and trajectory-based representations derived from the motion of traffic participants. Given a query scenario, each retrieval method produces a similarity score between the query and candidate scenarios. We evaluate both standalone retrieval performance and multimodal fusion of visual and trajectory-based similarity measures.

Each scenario is first preprocessed to extract relevant features (video frames, ego-motion matrices, and object trajectories), which are then encoded into a compact vector representation or distance matrix. Scenario similarity is expressed through cosine similarity or distance-based measures, depending on the modality. This section introduces the individual components of the framework.

\subsection{Visual Retrieval}
Visual retrieval methods represent a driving scenario using image or video embeddings extracted from RGB observations. These embeddings capture appearance, scene context, object semantics, and temporal visual cues. Similarity between scenarios is computed using cosine similarity in the embedding space. We evaluate two visual retrieval approaches.

\subsubsection{ViViT}
We employ ViViT \cite{vivit}, a transformer-based video model designed for capturing both spatial and temporal dependencies. The model tokenizes an input video into frame patches, processes them through multiple transformer layers, and produces a $d$-dimensional embedding vector. We use the hidden state of the first token, analogous to the [CLS] token in BERT \cite{ref_5}, as the global video representation. To improve computational efficiency, the videos are subsampled to 32 frames and cropped to exclude non-informative regions (e.g., sky or camera mount). All embeddings are precomputed offline, enabling fast similarity queries using cosine similarity:
$$
\cos(\mathbf{v}_i, \mathbf{v}_j) = \frac{\mathbf{v}_i \cdot \mathbf{v}_j}{\|\mathbf{v}_i\| \, \|\mathbf{v}_j\|}
$$
where $v_i$ and $v_j$ are the embeddings of two scenarios.

\subsubsection{Qwen3-VL}
We utilize Qwen3-VL-2B \cite{qwen} as a vision-language retrieval baseline due to its strong multimodal representation capabilities. For each driving scenario, 32 frames are sampled uniformly from the video clip and processed by the model to obtain a fixed-length embedding. Timestamp information is included in the visual input to preserve temporal context, enabling the model to distinguish between similar visual scenes occurring at different stages of a maneuver. We extract embeddings using the Matryoshka Representation Learning (MRL) representation with 768 dimensions and compute scenario similarity using cosine similarity between query and candidate embeddings. Unlike ViViT, which is trained specifically for video understanding, Qwen3-VL leverages large-scale multimodal pretraining and produces semantically rich representations that capture scene appearance, object context, and high-level traffic semantics.

\subsection{Trajectory-Based Retrieval}
While visual representations capture scene appearance and semantic context, many driving scenarios are primarily defined by the motion and interactions of traffic participants. We therefore investigate trajectory-based retrieval methods that compare scenarios using the temporal evolution of agent motion. We consider both explicit trajectory matching and learned trajectory representations.

\subsubsection{Exo-Trajectory-based Similarity Search}
Exo-Trajectory represents a scenario using the trajectories of surrounding traffic participants and measures similarity through explicit trajectory matching. Unlike visual embeddings, which rely on appearance and scene context, Exo-Trajectory directly compares motion patterns and interactions between agents.

To explicitly account for the behavior of other traffic participants, we developed the Exo-Trajectory similarity search method that compares the motion patterns of surrounding vehicles, pedestrians, and other dynamic objects. When 3D bounding boxes are available, we extract trajectories of all exo-objects and group them by type: vehicle, stopped vehicle, pedestrian, and rideable.

The trajectories are matched when the distance matrix is calculated. We used the Hungarian Algorithm \cite{ref_8} to find the best matching/assignment between trajectories. The Hungarian Algorithm is a combinatorial optimization algorithm that solves the assignment problem. To further refine the matching, we introduce a penalty mapping for unmatched trajectories on a per-class basis. In this mapping, unmatched vehicles incur a higher penalty, whereas, for example stopped vehicles incur a relatively lower penalty. This avoids overly penalizing scenarios that differ mainly by stationary objects, while emphasizing the importance of matching active ones. This solution ensures that scenarios with a similar number of objects and comparable movement patterns receive higher similarity values. 

The exo-trajectory-based similarity search process can be written as follows: 

\begin{itemize}
    \item \textbf{Distance matrix calculation for two scenarios.} An $N x M$ matrix is calculated using the Fréchet Distance between the trajectories of the two scenarios, where N and M are the number of trajectories of the first and second scenarios, respectively. The comparison disregards trajectories belonging to different groups (e.g., a vehicle and a pedestrian).
    \item \textbf{Matching similar trajectories from two scenarios.} The Hungarian Algorithm solves the assignment problem, resulting in an optimal matching between exo-trajectories. Unmatched exo-trajectories are penalized.
    \item \textbf{Aggregation.} The similarity between the two scenarios is calculated by aggregating the Fréchet-distance of matched trajectories while considering the penalties. A key factor here is the matched ratio, defined as \\
    $$matched = 1 - \frac{unmatched}{total}$$
\end{itemize}  
   
This rewards scenarios with more matched objects. This ratio is incorporated into a combined cost term along with the mean matched distance and the total unmatched penalty. Finally, the combined cost is converted into a [0,1] similarity measure through a scaling function:  

$$\frac{1}{1+\frac{combined}{k}}$$

The penalization process is especially beneficial when the two scenarios have similar numbers of other traffic participants, but these objects cannot be matched due to category mismatches. This way, the exo-trajectory-based similarity search ensures the sensitivity to the movement patterns of all objects, considering their object categories.

\subsubsection{ScenarioFormer}
ScenarioFormer is our proposed trajectory-based retrieval model. Unlike Exo-Trajectory, which relies on explicit trajectory matching, ScenarioFormer learns a compact embedding representation directly from agent trajectories using contrastive learning. The objective is to map scenarios exhibiting similar motion patterns to nearby locations in the embedding space while separating dissimilar scenarios.

While the Exo-Trajectory similarity search method provides an explainable solution for comparing autonomous driving scenarios, its high computation intensity might hinder large-scale deployment. Therefore, we developed a data-driven solution that enables a computationally less intensive precomputation method and allows fast querying. 

We designed, developed, and trained a model, called ScenarioFormer, which is based on the vision transformer model [9]. The exo-trajectories of a scenario can be represented by tensors with a shape $(N+1, T, C)$ where $N$ is the number of non-ego objects, $T$ is the number of timesteps, and $C$ is the number of attributes of the object. We set $N=25$ (with zero padding when the object number is less than 25), $T=15$, and $C=8$. The following bounding box properties were included in the input tensor after a BEV projection: 
\begin{itemize}
    \item longitudinal position (x coordinate),
    \item lateral position (y coordinate),
    \item object length,
    \item object width,
    \item heading (yaw angle),
    \item longitudinal velocity,
    \item lateral velocity.
    \item class ID
\end{itemize}

The input tensor is projected into a high-dimensional embedding space by a Linear layer. Then, positional encoding is added to retain the spatio-temporal information within the Transformer layers. The output of the model is a $d$-dimensional embedding.  

ScenarioFormer is a Siamese-Transformer \cite{ref_10} where the model is fed by two input tensors and is trained with contrastive learning. One input tensor represents the original scenario input, while the second input tensor is the augmented version of the first input. We utilized the following augmentations: 
\begin{itemize}
    \item random longitudinal offset,
    \item random lateral offset,
    \item random length extension,
    \item random width extension,
    \item random heading perturbation,
    \item random velocity noise,
    \item random dropout of objects.
\end{itemize}

\begin{figure*}[!t]
    \centering

    \begin{subfigure}{\linewidth}
        \centering
        \includegraphics[width=\textwidth,height=0.45\textheight,keepaspectratio]{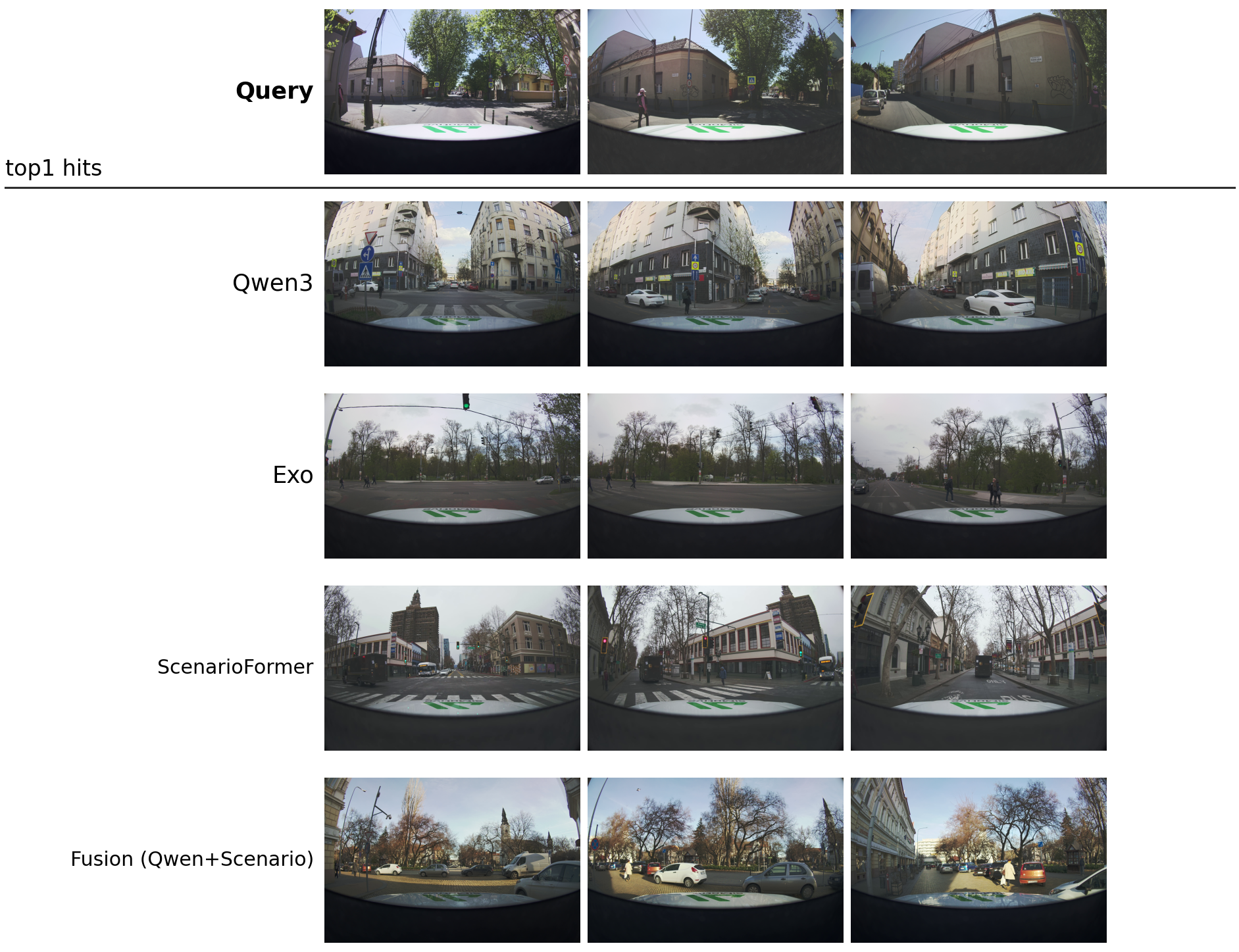}
        \caption{Pedestrian crossing scenario}
        \label{fig:pedestrian}
    \end{subfigure}

    \vspace{2mm}

    \begin{subfigure}{\linewidth}
        \centering
        \includegraphics[width=\textwidth,height=0.45\textheight,keepaspectratio]{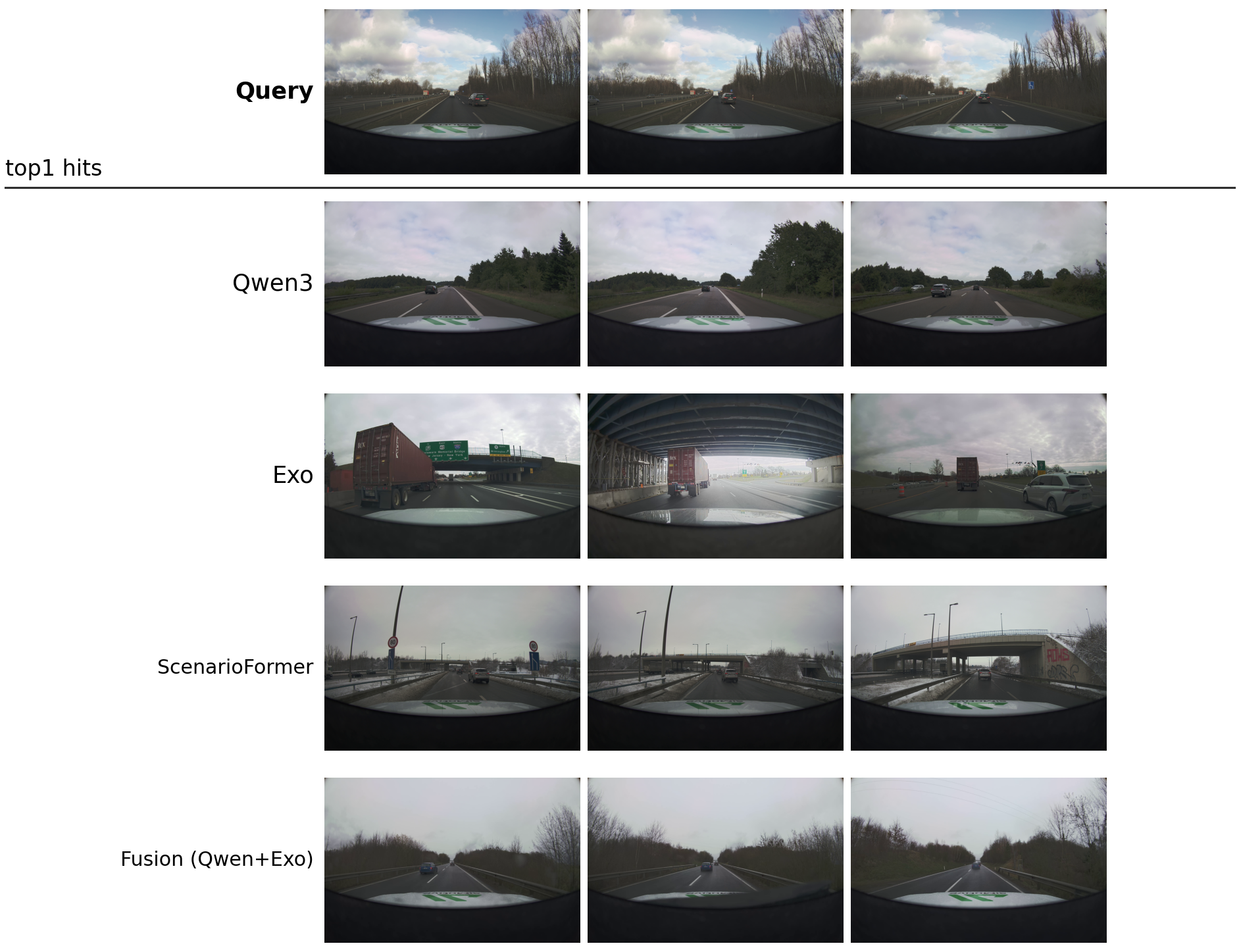}
        \caption{Cut-in scenario}
        \label{fig:cutin}
    \end{subfigure}

    \caption{Examples from the evaluation benchmark, illustrating representative pedestrian-crossing and cut-in scenarios. Best viewed by zooming in.}
    \label{fig:benchmark_examples}
\end{figure*}

Contrastive learning aims to pull similar scenarios close to each other in the embedding space while pushing different scenarios far away. Similar scenarios can be generated by augmenting the original scenario with the above-mentioned augmentations. These (original, augmented) scenarios are the positive pairs that will be close to each other in the embedding space. Negative pairs can easily be formed by taking all other scenarios contained by the same batch. The metric used to determine the similarity between two scenarios is cosine similarity. We used the InfoNCE loss represented by a CrossEntropy loss. Since contrastive learning is sensitive to batch size, we utilized gradient accumulation to increase the effective batch size without running out of GPU memory. 

The scalability of this approach is ensured by the precomputation of the embeddings. Then, the same similarity search described in the video-based method can be used. To find similar scenarios to a new one, it has to be encoded by the model, and then cosine similarity can be used to query similar scenarios from the embedding pool. 

\subsection{Multimodal Fusion}

Visual and trajectory representations capture complementary aspects of scenario similarity. To combine their strengths, we perform score-level fusion between the visual and trajectory retrieval modules. Given a visual similarity score $s_{\text{vision}}$ and a trajectory similarity score $s_{\text{traj}}$, the final similarity score is computed as

\begin{equation}
s = \alpha s_{\text{traj}} + (1-\alpha)s_{\text{vision}},
\end{equation}

where $\alpha$ controls the contribution of the trajectory representation. Fusion allows the retrieval system to leverage both appearance-based and motion-based cues when ranking candidate scenarios.

\section{Experiments}
\label{sec:experiments}
\subsection{Benchmark Dataset and Evaluation Protocol}

We evaluate our scenario retrieval solution on a subset of the extended version of the aiMotive Multimodal Dataset \cite{aimotive}. The benchmark contains 13 scenario categories covering both dynamic traffic interactions and appearance-driven scene characteristics. Following our focus on motion-aware retrieval, we primarily report results on a challenging dynamic subset consisting of seven topics: exo-vehicle cut-ins, highway cut-ins, pedestrian crossings, pedestrian crossings during ego turns, cyclist interactions, ego turning maneuvers, and traffic queueing.

Each topic contains approximately 100 manually annotated clips with graded similarity labels (overall annotated dataset size is 682). Clips are assigned relevance scores from 0 to 3, where 3 denotes highly similar scenarios, 2 denotes similar scenarios with minor differences, 1 denotes weak similarity, and 0 denotes unrelated scenarios. The similarity labels were manually assigned by domain experts familiar with autonomous-driving scenarios. Every grade-3 clip (overall 97) serves as a query and is evaluated against all the clips.

Our primary metric is Normalized Discounted Cumulative Gain (NDCG@10), which accounts for graded relevance and rewards rankings that place highly similar scenarios above partially matching ones. We additionally report Recall (Recall@k$_{\ge g2}$: fraction of clearly-useful clips retrieved in top-k, Recall@k$_{(g3)}$: fraction of strict positives found in top-15) metric. All methods are evaluated using a unified candidate pool and identical query sets, enabling direct comparison between visual and trajectory-based approaches.

\begin{table}[t]
\centering
\caption{Standalone retrieval performance on the dynamic subset of the benchmark.
Higher is better for all metrics.}
\label{tab:standalone_results}
\begin{tabular}{lcccc}
\toprule
Method & NDCG@10 & Rec@15$_{\ge2}$ & Rec@15$_{(3)}$ \\ %& MAP@10 \\
\midrule
ViViT & 0.514 & 0.421 & 0.363 \\ %& 0.169 \\
ScenarioFormer & 0.565 & 0.441  & 0.388 \\ % & 0.207 \\
Exo-Trajectory & 0.565 & 0.475 & 0.434 \\ % & 0.209 \\
Qwen3-VL-2B & \textbf{0.621} & \textbf{0.513} & \textbf{0.463} \\ % & \textbf{0.241} \\
\bottomrule
\end{tabular}
\end{table}

\subsection{Standalone Retrieval Performance}

Table~\ref{tab:standalone_results} compares the four retrieval modules evaluated in this work. Two methods operate on visual representations (ViViT and Qwen3VL-2B), while two methods rely exclusively on trajectory information (Exo-Trajectory and ScenarioFormer).

Qwen3-VL-2B achieves the strongest overall performance, reaching an NDCG@10 of 0.621. Among the trajectory-based approaches, Exo-Trajectory and ScenarioFormer obtain nearly identical performance (both 0.565 NDCG@10), substantially outperforming the weaker visual baseline (ViViT, 0.514).

The per-topic analysis reveals a complementary behavior between visual and trajectory representations. Visual embeddings perform best on scenarios where appearance cues and semantic context provide strong signals, including pedestrian interactions and cyclist-related events. In contrast, trajectory-based methods excel in scenarios whose defining characteristics arise primarily from motion patterns. Traffic queueing and turning maneuvers are particularly well captured by trajectory representations, indicating that explicit motion modeling provides information that is difficult to infer from appearance alone.

These results suggest that visual and trajectory modalities capture different aspects of scenario similarity and motivate their combination through multimodal fusion.

\begin{table}[t]
\centering
\caption{Best score-level fusion results on the dynamic benchmark. $\Delta$ denotes the improvement in NDCG@10 over the corresponding vision-only baseline.}
\label{tab:fusion_results}
\begin{tabular}{lccc}
\toprule
Fusion Method & $\alpha$ & NDCG@10 & $\Delta$ \\
\midrule
ViViT + ScenarioFormer & 0.4 & 0.599 & +0.085 \\
ViViT + Exo-Trajectory & 0.5 & 0.616 & +0.102 \\
Qwen3 + ScenarioFormer & 0.3 & 0.662 & +0.041 \\
Qwen3 + Exo-Trajectory & 0.3 & \textbf{0.671} & +0.050 \\
\bottomrule
\end{tabular}
\end{table}

\subsection{Multimodal Fusion}

To combine visual and trajectory information, we perform score-level fusion,

\begin{equation}
s = \alpha s_{\text{traj}} + (1-\alpha)s_{\text{vision}},
\end{equation}

where $\alpha$ controls the contribution of the trajectory model.

Results are summarized in Table~\ref{tab:fusion_results}. Fusion consistently improves retrieval performance over the corresponding visual baseline, demonstrating that trajectory information contributes complementary cues not captured by appearance embeddings.

The strongest overall result is obtained by combining Qwen3-VL-2B with Exo-Trajectory, yielding an NDCG@10 of 0.671, corresponding to an absolute improvement of 5.0\% over Qwen3-VL-2B alone. Similarly, combining Qwen3-VL-2B with ScenarioFormer increases NDCG@10 from 0.621 to 0.662. The effect is even larger for the weaker ViViT baseline, where fusion with Exo-Trajectory improves NDCG@10 from 0.514 to 0.616.

The per-topic analysis in Table~\ref{tab:fusion_per_topic}
reveals that the benefit of fusion depends strongly on the
scenario category. Trajectory information is particularly
valuable for cut-in and cyclist-interaction scenarios, where the Exo-Trajectory fusion achieves the best performance. In contrast, pedestrian-related scenarios and traffic queueing are best handled by the ScenarioFormer-based fusion, suggesting that the learned trajectory embedding captures higher-level semantic interactions between traffic participants. Overall, no single fusion strategy dominates across all topics, further supporting the hypothesis that visual and trajectory representations capture complementary aspects of scenario similarity.

Overall, the fusion experiments demonstrate that motion-aware retrieval and visual retrieval are complementary and that combining both modalities yields the most robust similarity search system. 

\begin{table}[t]
\centering

\caption{Per-topic NDCG@10 comparison between Exo-Trajectory and ScenarioFormer. Positive $\Delta$ indicates an advantage for Exo-Trajectory, while negative values favor ScenarioFormer. Although both methods achieve nearly identical macro performance, they exhibit
complementary strengths. Exo-Trajectory performs better on cut-in and cyclist scenarios, while ScenarioFormer excels on pedestrian interactions and traffic queueing.}
\label{tab:trajectory_comparison}
\begin{tabular}{lccc}
\toprule
Topic & Exo & Scene & $\Delta$ \\
\midrule
01 Cut-in (env. invariant) & \textbf{0.491} & 0.377 & +0.114 \\
02 Highway cut-in & \textbf{0.597} & 0.473 & +0.124 \\
03 Pedestrian crossing & 0.613 & \textbf{0.703} & -0.090 \\
04 Pedestrian + ego turn & 0.417 & \textbf{0.564} & -0.147 \\
05 Cyclist interaction & \textbf{0.525} & 0.465 & +0.060 \\
06 Ego turning at intersection & \textbf{0.511} & 0.489 & +0.022 \\
07b Traffic queue & 0.802 & \textbf{0.884} & -0.082 \\
\midrule
Macro Average & \textbf{0.565} & \textbf{0.565} & +0.000 \\
\bottomrule
\end{tabular}
\end{table}

\begin{table*}[t]
\centering
\caption{Best per-topic NDCG@10 obtained by each fusion method. For each topic,
the fusion weight $\alpha$ is independently selected to maximize NDCG@10.
Bold values indicate the best fusion result for that topic.}
\label{tab:fusion_per_topic}
\begin{tabular}{lcccc}
\toprule
Topic & ViViT+ScenarioFormer & ViViT+Exo & Qwen3-VL+ScenarioFormer & Qwen3-VL+Exo \\
\midrule
01 Cut-in (env. invariant)               & 0.397 & \textbf{0.491} & 0.405 & \textbf{0.491} \\
02 Highway cut-in        & 0.612 & \textbf{0.682} & 0.658 & 0.680 \\
03 Pedestrian crossing   & 0.793 & 0.777 & \textbf{0.869} & 0.854 \\
04 Pedestrian + ego turn & 0.617 & 0.604 & \textbf{0.713} & 0.691 \\
05 Cyclist interaction   & 0.473 & 0.537 & 0.649 & \textbf{0.663} \\
06 Ego turning at intersection & \textbf{0.575} & 0.523 & 0.549 & 0.531 \\
07b Traffic queue        & 0.890 & 0.843 & \textbf{0.920} & 0.913 \\
\bottomrule
\end{tabular}
\end{table*}

\subsection{Analysis of Trajectory Representations}

We next compare the two trajectory representations directly. Although their overall NDCG@10 scores are nearly identical, they exhibit different strengths.

Exo-Trajectory performs best on cut-in scenarios and cyclist interactions, achieving substantial gains on both highway and urban cut-in topics. These scenarios are characterized by distinctive geometric motion patterns, which are effectively captured by explicit trajectory matching using Fréchet-distance-based costs.

In contrast, ScenarioFormer performs best on pedestrian-centric scenarios and traffic queueing. The learned representation benefits from object-type embeddings and semantic augmentations, enabling it to better distinguish interactions involving different actor categories.

The complementary behavior of the two trajectory models explains why both serve as effective fusion partners despite having similar aggregate performance. Exo-Trajectory provides stronger geometric matching, while ScenarioFormer captures higher-level semantic relationships between dynamic actors.

\begin{table}[t]
\centering
\caption{Ablation study of the ScenarioFormer. Results are reported as macro
NDCG@10 on the dynamic benchmark.}
\label{tab:ablation}
\begin{tabular}{lcc}
\toprule
Configuration & NDCG@10 & Gain \\
\midrule
Simple (global) augm. & 0.444 & -- \\
Hard (object-level) augm. & 0.510 & +0.066 \\
Semantic (class-aware) augm. & 0.516 & +0.006 \\
Semantic augm. + 2$\times$ data & 0.531 & +0.015 \\
+ Ego-motion augm. & \textbf{0.565} & +0.034 \\
\bottomrule
\end{tabular}
\end{table}

\subsection{Ablation Study}

Table~\ref{tab:ablation} presents an ablation study of the ScenarioFormer training procedure. We evaluate progressively stronger augmentation strategies while keeping the model architecture fixed.

Starting from a simple augmentation policy (global jitter), introducing stronger trajectory perturbations (per-object jitter, dropout, position noise) increases NDCG@10 from 0.444 to 0.510. Adding class-aware semantic augmentations (class-aware jitter, dropout) yields a further improvement to 0.516, while increasing the amount of training data (from 150k sequences to over 300k) raises performance to 0.531.

The final improvement comes from incorporating ego-vehicle
motion into the scenario representation. Incorporating an ego-motion together with maneuver-preserving perturbations increases NDCG@10 to 0.565, producing the best-performing ScenarioFormer model.

The ablation study highlights augmentation design as the dominant factor affecting retrieval quality. Weak augmentations allow the contrastive objective to exploit trivial correspondences between positive pairs, whereas stronger semantic and motion-preserving augmentations encourage learning representations that capture meaningful scene-level structure.

\subsection{Discussion}

The experimental results reveal three key findings. First, modern vision-language embeddings provide a strong baseline for scenario retrieval, particularly when semantic appearance cues are informative. Second, trajectory representations remain highly valuable for events defined primarily by motion dynamics, such as cut-ins, turning maneuvers, and traffic queueing. Third, combining visual and trajectory information consistently yields the best overall performance, demonstrating that the two modalities encode complementary notions of similarity.

\begin{figure}[t]
    \centering
    \includegraphics[width=\textwidth,height=0.9\textheight,keepaspectratio]{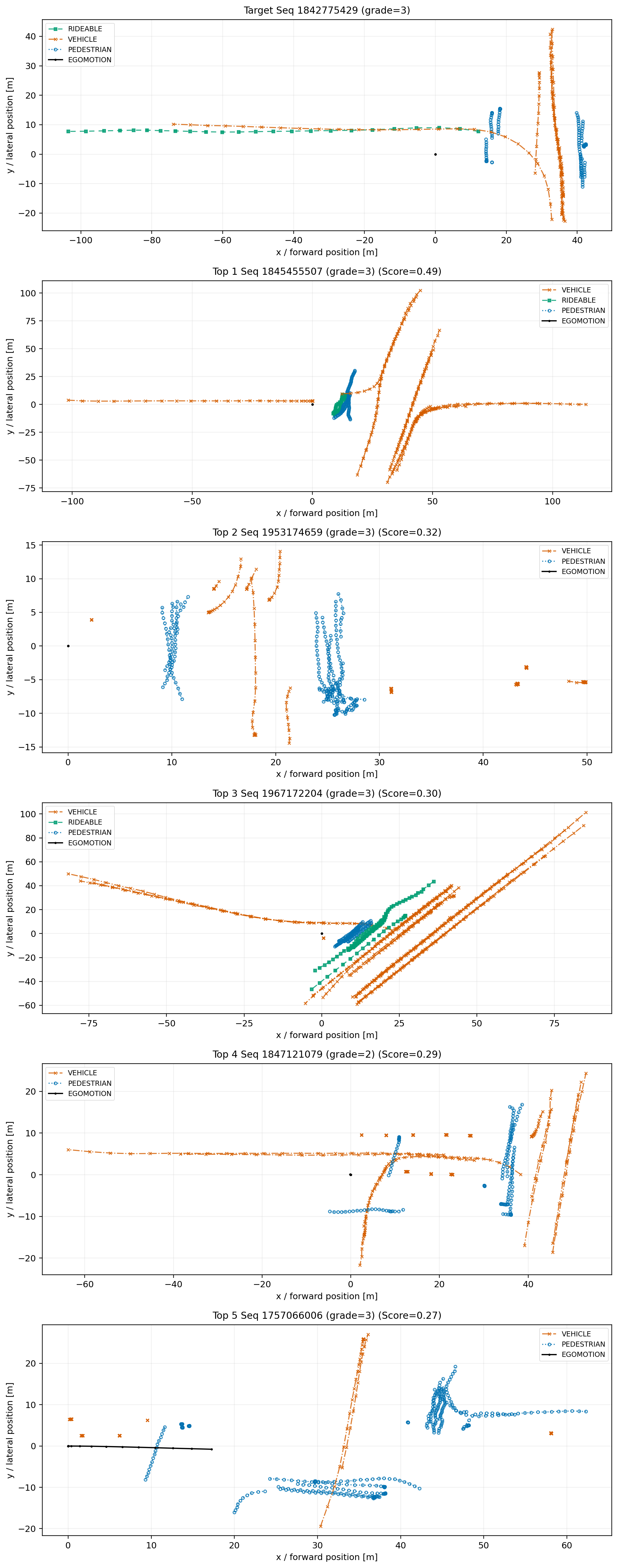}
    \caption{Visualization of the top-5 retrieval result of ScenarioFormer in a pedestrian crossing use case.}
    \label{fig:scenario_top_5}
\end{figure}

These findings support the use of multimodal retrieval systems for autonomous-driving data mining, where relevant scenarios often depend simultaneously on visual context and temporal motion patterns. Figures~\ref{fig:benchmark_examples} and~\ref{fig:scenario_top_5} provide qualitative examples of the benchmark and retrieval results. Figure~\ref{fig:benchmark_examples} depicts query scenarios together with the top-1 retrieval returned by each model, highlighting the different notions of similarity captured by visual and trajectory-based representations. Figure~\ref{fig:scenario_top_5} presents the top-5 retrieval results obtained by ScenarioFormer for a pedestrian-crossing query, demonstrating that the learned trajectory representation successfully retrieves scenarios exhibiting similar pedestrian-vehicle interactions despite variations in visual appearance and scene context.

\section{Conclusion}

We presented a multimodal framework for autonomous-driving scenario similarity search that combines visual and trajectory-based representations. We evaluated two off-the-shelf vision models and two proposed trajectory-based approaches on a manually annotated driving-scene similarity benchmark. The results show that trajectory information provides a strong retrieval signal for motion-centric scenarios such as cut-ins, turning maneuvers, and traffic queueing, while visual embeddings perform best when appearance cues are informative. Although the strongest vision model achieved the best standalone performance, both trajectory-based methods achieved competitive results and exhibited complementary strengths. Most importantly, combining visual and trajectory information consistently improved retrieval quality, yielding the best overall performance.

Future work will focus on larger-scale benchmarks, end-to-end multimodal representations that jointly encode visual context and agent motion, and retrieval-driven applications for dataset curation and autonomous-driving system validation.

{
    \small
    \bibliographystyle{ieeenat_fullname}
    \bibliography{main}
}

\end{document}